\documentclass[runningheads]{llncs}
\usepackage{graphicx}
\graphicspath{images}
\usepackage{array}
\newcolumntype{P}[1]{>{\centering\arraybackslash}p{#1}}

\begin{document}
\title{Attention based Multiple Instance Learning for Classification of Blood Cell Disorders}
\titlerunning{MIL Classification of Blood Disorders}

\author{Ario Sadafi\inst{1,2,3} \and
Asya Makhro\inst{4} \and
Anna Bogdanova\inst{4} \and
Nassir Navab \inst{2,5} \and \\
Tingying Peng	 \inst{1,2\star} \and
Shadi Albarqouni \inst{2,6\star} \and
Carsten Marr \inst{1}\thanks{Shared senior authorship}}
\authorrunning{A. Sadafi et al.}
\institute{Institute of Computational Biology, Helmholtz Zentrum München - German Research Center
for Environmental Health, Germany \and
Computer Aided Medical Procedures, Technical University of Munich, Germany
\and
Helmholtz AI, Helmholtz Center Munich, Germany
\and
Red Blood Cell Research Group, Institute of Veterinary Physiology, Vetsuisse Faculty and the Zurich Center for Integrative Human Physiology, University of Zurich, Zurich, Switzerland
\and
Computer Aided Medical Procedures, Johns Hopkins University, USA
\and
Computer Vision Lab (CVL), ETH Zurich, Switzerland}
\maketitle              
\begin{abstract}
Red blood cells are highly deformable and present in various shapes. 
In blood cell disorders, only a subset of all cells is morphologically altered and relevant for the diagnosis. However, manually labeling of all cells is laborious, complicated and introduces inter-expert variability.
We propose an attention based multiple instance learning method to classify blood samples of patients suffering from blood cell disorders. 
Cells are detected using an R-CNN architecture. With the features extracted for each cell, a multiple instance learning method classifies patient samples into one out of four blood cell disorders. 
The attention mechanism provides a measure of the contribution of each cell to the overall classification and significantly improves the network’s classification accuracy as well as its interpretability for the medical expert.

\end{abstract}
\section{Introduction}

Historically, classification of hereditary hemolytic anemias, a particular class of blood disorders, is based on the abnormal shape of red blood cells. “Sickle cell disease”, “spherocytosis”, “ovalocytosis”, “stomatocytosis”: these types of anemia refer directly to the changes in cell shape, whereas genetic causes of the disease were identified later \cite{gallagher2004update,gallagher2005red,kato2018sickle}. All but one (sickle cell disease) of the above-mentioned disorders are “structural diseases” of the red blood cell membrane caused by mutations in genes coding for the cytoskeletal proteins spectrin, ankyrin and band 3 protein, as well as protein 4.2 \cite{gallagher2005red} making cells look like flowers, stars, hedgehogs, cups, droplets or spheres
\cite{bessis1974corpuscles}. Characteristic changes in morphology are hallmarks of diseases caused by abnormalities in hemoglobin structure as for sickle cell disease and thalassemia. Somewhat more subtle are the changes in shape of red blood cells of patients harboring mutated glycolytic enzymes \cite{grace2018red} or ion channels (like the Gardos channel or the PIEZO1 channel \cite{picard2019clinical}). Independent of the cause and class of hereditary anemia, not all red blood cells but a fraction of them (often as large as 5-10\% of the total cell population) are abnormally shaped. This makes diagnosis based on shape changes alone difficult, and additional tests are currently required. Furthermore, detection of abnormal shapes suffers from the subjective view of a human observer, a skillful, but possibly biased expert that may only process several hundreds of cells per patient.
Instead, machine learning approaches are required and have been showing to outperform human experts in a number of clinical tasks. Classification of skin cancer at the dermatologist level proposed by Esteva et al. \cite{esteva2017dermatologist}, human level recognition of blast cells by Matek et al. \cite{matek2019human} or an AI system for breast cancer screening developed by McKinney et al. \cite{mckinney2020international} are just some of the various cases machine learning excels experts.
Introduction of an unbiased computer-based assessment of red blood cell shapes and their abundance in a blood sample may open new possibilities for diagnostics, assessment of disease severity, and monitoring of treatment success for the patients with rare anemias.


Multiple instance learning (MIL) is used in medical image computation when all of the instances from a patient must be taken into account and no specific label exists for each of the instances. For example, Campanella et al. \cite{campanella2019clinical} propose a method to whole slide pathology image classification with the MIL. Here, each whole slide image is weakly labeled as healthy or cancerous, but no specific label exists for every small image patch. Similarly, Ozdemir et al. \cite{ozdemir20193d}  suggest a method based on MIL to classify lung cancer CT scans. In a slightly different work, Conjeti et al. \cite{conjeti2017deep} suggest a method of hashing for medical image retrieval based on MIL and the auxiliary branch for the vanishing gradient problem known to impede MIL approaches. While these approaches perform on a patch level, none of them is able to identify single cells that are often crucial for the diagnosis.

To this end, we propose a method based on weak patient labels and attention based MIL to classify patient blood samples into disease classes. Cells are extracted from images by an R-CNN architecture previously trained on a single cell detection task and feature maps of the backbone ResNet are passed to the proposed method. Without any cellwise labeling, the model manages to detect landmark cells for every disease in an unsupervised way by giving them the highest attention.


\section{Methodology}
\begin{figure}[t]
    \centering
    \includegraphics[width=1\textwidth, page=1, trim=0 4.8cm 0 0, clip]{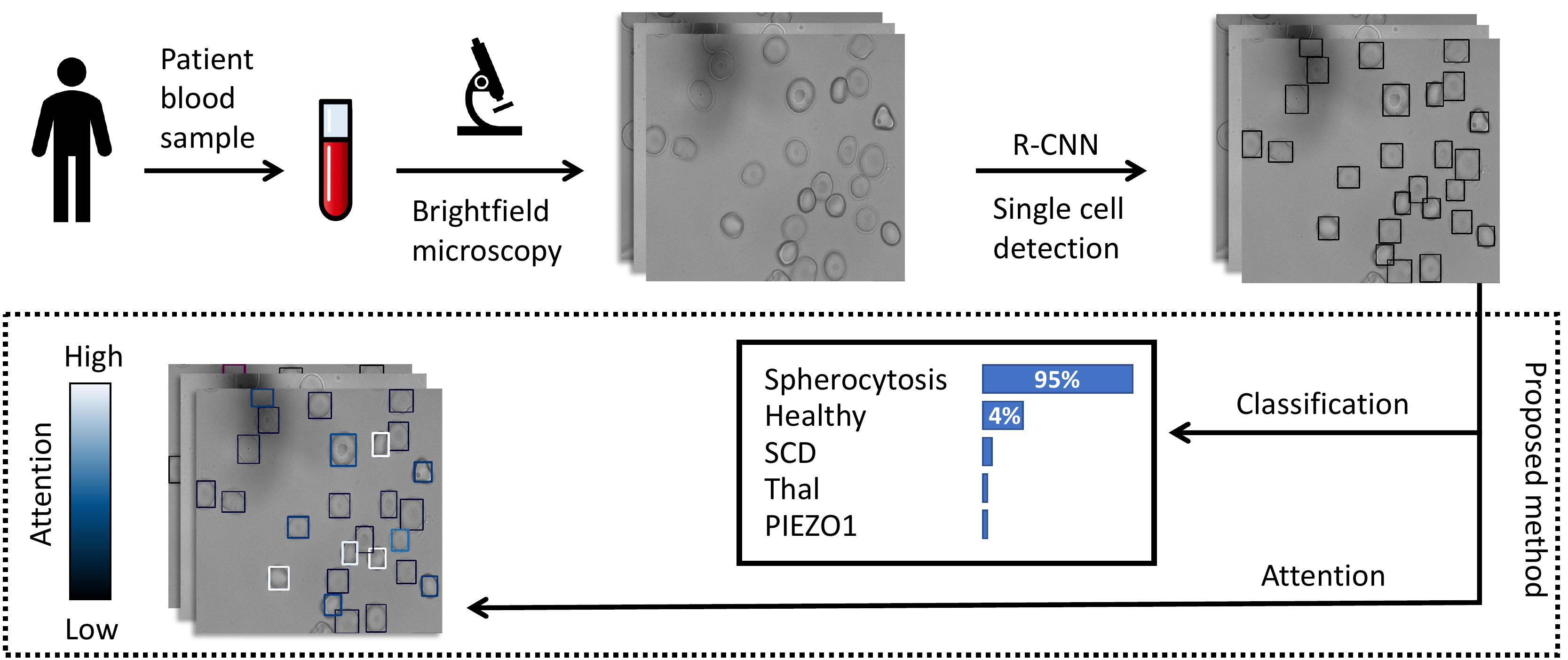}
    \caption{Overview of the proposed method. Bright-field images of blood samples are acquired with a microscope. Using a previously trained R-CNN, all cells are detected from each image. Looking at all of the cells our proposed approach classifies the sample and provides a cell-wise attention score for better interpretability.}
    \label{overviewfigure}
\end{figure}
\begin{figure}
    \centering
    \includegraphics[width=0.9\textwidth, page=2, trim=0 7.3cm 0 0, clip ]{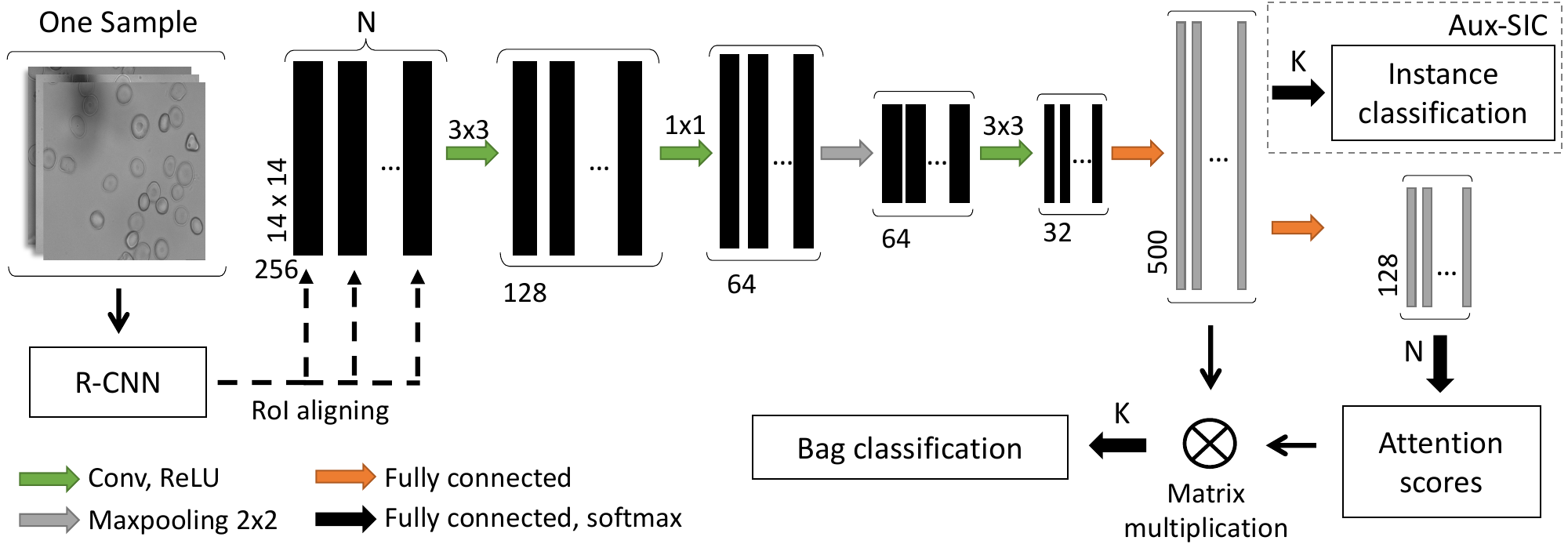}
    \caption{Architecture of the proposed method. The R-CNN extracts features from the input images and detects the red blood cells. Based on the detected cells a bag of cell features in all of the images is formed after RoI aligning. Passing through convolutional and fully connected layer and attention pooling a feature vector is shaped for the bag and classified. An auxiliary single instance classification (Aux-SIC) branch helps the training during the first steps. N is the number of instances in a bag and K the total number of classes.}
    \label{arcitecturefigure}
\end{figure}

A patient’s blood sample may consist of several bright-field images (see Dataset for a detailed description) and each image contains several instances / single red blood cells (see Fig. \ref{overviewfigure}). 
A previously trained R-CNN architecture is used to find instances and extract their features (see Fig. \ref{arcitecturefigure}). 
Based on these features, we propose to classify a sample into one of four diseases taking into account the features of all of the instances present in the input. 
Additionally, an attention score improves performance and interpretability of the method to medical experts for further verification.
The proposed approach consists of three main blocks: 
(i) the multiple instance learning, 
(ii) an auxiliary single instance classifier and 
(iii) the attention mechanism 
(see Fig. \ref{overviewfigure}).

More formally, our objective is a model $\mathrm{f}$ that classifies a blood sample containing several instances into one of the classes $\mathrm{c_i \in C}$ and generates a score $\mathrm{\vec{a_k} \in A}$ denoting the contribution of each instance in the final decision:

\begin{equation}
    \mathrm{c_i,\vec{a_k} = f(B)},
\end{equation}

where $\mathrm{B=\{I_1,...,I_N\}}$ is the bag of instances and each instance $\mathrm{I_i}$ is a tensor of the size $256\times14\times14$.

\subsection{Preprocessing: Single cell detection}
Any off the shelf detection algorithm can be used for detecting single cells in the images. 
We are using a modified Mask R-CNN architecture \cite{he2017mask} with ResNet\cite{he2016deep} backbone that generates the features. For every detected cell the relevant features are extracted with RoI aligning and used as an instance $\mathrm{I}$ given as the input to our method.
The Mask R-CNN was trained on a separate dataset from \cite{sadafi2019multiclass}, consisting of 208 microscopic images with each containing 30-40 red blood cells (total of 
$>7000$ single cells) annotated by a biomedical expert. The Mask R-CNN achieves a mean average precision (mAP) of 0.91 which is accurate enough for our application.
In order to prevent cumbersome cell by cell annotation for segmentation which is not required in this approach, we limited the Mask R-CNN to a binary classification and disabled the mask head.
This way, annotation of data for single cell extraction is limited to bounding boxes around the cells that does not require any special expertise and can be performed by anyone with minimum training.


\subsection{Single instance classification}
Single instance classification (SIC) is the most intuitive approach for classification of samples. 
In our case, each instance is passed through several convolutional layers and a embedding feature vector $\vec{h}$ for every instance in the bag is generated (Fig. \ref{arcitecturefigure}).
SIC is a CNN architecture that classifies each instance embedding based on the weak labels of the bag. At inference time, a majority voting is employed to determine the class of a given sample based on the single cells' classification results:

\begin{equation}
\mathrm{
    \mathcal{L}_{SIC}(\theta, \psi) = \frac{1}{N}\sum_{i=1}^{N} \mathrm{CE}(\vec{c}_i, \hat{\vec{c}}_i), 
    }
\end{equation}
where 
$\mathrm{\hat{\vec{c}}_i = f_{SIC}(\vec{h}_i; \psi)}$, $\mathrm{CE}$ is the cross entropy loss, and 
\begin{equation}
    \mathrm{\vec{h}_i = f_{EMB}(I_i, \theta) : I_i \in B}
    \label{eqH}
\end{equation}
with $\vec{c}_i$ being the label for each cell $i$ based on the bag label. $\psi$ and $\theta$ are learned model parameters.
\subsection{Multiple instance learning}
In contrast to supervised methods where for every given instance one tries to find a target variable $\mathrm{\hat{\vec{c}} \in C}$, multiple instance learning (MIL) tries to find the target variable $\mathrm{\hat{\vec{c}}}$ based the input which is a set of instances $\mathrm{B = \{I_1, ..., I_N\}}$.
There are two approaches to implement MIL: instance level and embedding level approaches.
We use the embedding level approach to formulate the MIL problem. 
As defined in eq. \ref{eqH}, the function $\mathrm{f_{EMB}}$ maps every instance into a low dimensional space $\mathrm{\vec{h}}$ and a single representation for the whole bag $\mathrm{z}$ is generated using a MIL pooling method. A bag level classifier classifies $\mathrm{z}$ into one of the classes.
A few MIL pooling methods exist. 
One popular method is max pooling \cite{amores2013multiple} where maximization is used to generate the bag level representation:
\begin{equation}
    \mathrm{z_m = \max_{k=1...N}\{h_{km}\}.}
\end{equation}

The MIL approach can be formulated as follows:

\begin{equation}
    \mathrm{\mathcal{L}_{MIL}(\theta, \phi) = CE(\vec{c}, \hat{\vec{c}}), }
\end{equation}
where $\mathrm{\hat{\vec{c}} = f_{MIL}(\{\vec{h}_1,\ldots,\vec{h}_N\}, \{\alpha_1, \ldots,\alpha_N\}; \phi)}$, where $\alpha$ is the attention score (see Sec.~\ref{sec:attention}), and $\phi$ represents learned model parameters.

\subsection{Attention mechanism}
\label{sec:attention}
An attention mechanism is widely used in various deep learning tasks from semantic segmentation \cite{Chen_2016_CVPR} to conversational question answering \cite{zhu2018sdnet}.
Ilse et al. \cite{ilse2018attention} proposed an attention mechanism where a weighted average is calculated over the instance embeddings and these weights are learned by the neural network.
If $\mathrm{H=\{h_1, ... , h_N\}}$ is a bag of instance embeddings, attention based MIL pooling is defined as:
\begin{equation}
    \mathrm{z = \sum^N_{k=1} \alpha_kh_k,}
\end{equation}
where
\begin{equation}
    \mathrm{\alpha_k = \frac{exp\{w^Ttanh(Vh_k^T)\}}{\sum_{j=1}^N exp\{w^Ttanh(Vh_j^T)\}}.}
\end{equation}

$V$ and $w$ are parameters that are learned during the training. This attention scores $\mathrm{\alpha_k}$ help the interpretability of the trained model by discovering the contribution of each instance in the drawn conclusion and acting as a similarity measure for comparison between the instances.

\subsection{Overall objective function}
One of the difficulties of MIL is sparse and vanishing gradients due to instance pooling. 
Here, we propose a dynamic loss function that incorporates the MIL loss along with the auxiliary SIC branch during the training using a decaying coefficient defined as follows:

\begin{equation}
    \mathrm{\mathcal{L}(\theta, \phi, \psi) = (1 - \beta^{E})\mathcal{L}_{MIL} + \beta^{E} \mathcal{L}_{SIC}, }
\end{equation}
where $\beta$ is a hyper-parameter, and $\mathrm{E}$ is the epoch index. 

\section{Experiments}
We validated the proposed method on a dataset of bright-field microscopy images of human blood cell genetic disorders. 
We designed an ablation study as follows:
(i) single instance classification (SIC), 
(ii) multiple instance learning (MIL) with maxpooling, 
(iii) MIL with maxpooling and the auxiliary SIC branch, and (iv) MIL with attention pooling and auxiliary SIC branch.

\subsubsection{Dataset.}
All images are obtained by an Axiocam mounted on Axiovert 200m Zeiss microscope with a 100x objective. 
No preprocessing is done and cells are not stained. 
The data consists of patient samples acquired at different time points, or in different kinds of solutions. 
In each sample there are 4 - 12 images and each image contains 12 - 45 cells.
The dataset contains four genetic morphological disorders: Thalassemia (3 patients, 25 samples), sickle cell disease (9 patients, 56 samples), PIEZO1 mutation (8 patients, 44 samples) and hereditary spherocytosis (13 patients, 89 samples). Also we have a healthy control group (26 individuals, 137 samples). 
We did patient-wise train and test split for a fair test set selection.

Patients previously diagnosed with hereditary spherocytosis were enrolled in the CoMMiTMenT-study (\url{http://www.rare-anaemia.eu/}). 
This study was approved by the Medical Ethical Research Board of the University Medical Center Utrecht, the Netherlands, under reference code 15/426M and by the Ethical Committee of Clinical Investigations of Hospital Clinic, Spain (IDIBAPS) under reference code 2013/8436.
Genetic cause of the disease was identified by the research group of Richard van Wijk, University Medical Center Utrecht, the Netherlands.
The healthy subjects study was approved by the ethics committees of the University of Heidelberg, Germany, (S-066/2018) and of the University of Berne, Switzerland (2018-01766), and was performed according to the Declaration of Helsinki.

\subsubsection{Implementation Details.}
The proposed method consists of three components: multiple instance embedding, auxiliary SIC, attention \& bag classifier. Fig. \ref{arcitecturefigure} shows the architecture of the method.
\paragraph{The multiple instance embedding} is a multi-layer convolutional neural network used for embedding features extracted by the R-CNN. It consists of five convolutional layers, a dropout layer with probability of $0.1$ after the first convolution, a maxpooling layer and a fully connected layer that creates the representative 1-D embedding feature vector. 
These layers are common between both SIC and MIL branches and remain trainable by both branches.
\paragraph{An auxiliary single instance classifier} looks at every instance embedding and tries to classify it with a fully connected layer. 
We chose a $\beta$ equal to $0.5$ to have a decaying contribution of this auxiliary branch during the training. Starting with a high contribution at the beginning and gradually reaching zero towards the end of the training.

\paragraph{In the attention and bag classifier}
the matrix of embedded instance representations ($H$) is multiplied by the attention matrix. The attention matrix is dynamically generated based on $H$. After the multiplication bag classifier, a fully connected layer with softmax, does the final MIL classification.

\paragraph{Training.}
 We decided to use 3-fold cross validation. Three different, independent models are trained based on each fold and performances are averaged. The models are trained by AMSGrad variation of Adam optimizer \cite{amsgrad} with a learning rate of $0.0005$ and the weight decay coefficient of $10^{-5}$. 
Training continues for a maximum of 150 epochs with an early stopping if the MIL loss drops below a specific threshold ($0.005$) for five consecutive epochs.
The same training parameters are used across all conducted experiments. Further details about hyper parameters and implementation details can be found in our repository under \url{https://github.com/marrlab/attMIL}.

\paragraph{Evaluation metrics.}
 Accuracy, macro F1 score and average area under the ROC and precision recall curves are used for comparison between different approaches.

\paragraph{Baselines.}
SIC is the baseline for our approach. 
We compare the results with a MIL without the designed auxiliary SIC branch and a MIL with a maxpooling method to our approach.

\begin{table}[t]
\caption{Comparison of the proposed method (MIL + att. + SIC) with other baselines. Mean and standard deviation for accuracy, weighted F1 score and average of area under ROC curve of all classes for five runs is shown.}
\label{tableresults}
\begin{center}
\begin{tabular}{p{3cm}|P{2.5cm}|P{2.5cm}|P{2.5cm}}
\textbf{Method} & \textbf{Accuracy} & \textbf{F1 Score} & \textbf{AU ROC} \\\hline
SIC & $0.50 \pm 0.01$ & $0.46 \pm 0.01$ & $0.743 \pm 0.005$    \\\hline
MIL + max & $0.46 \pm 0.04$ & $0.33 \pm 0.05$ & $0.644 \pm 0.049$  \\\hline
MIL + max + SIC & $0.70 \pm 0.01$ & $0.69 \pm 0.11$ & $0.916 \pm 0.005$  \\\hline
MIL + att. + SIC & $\mathbf{0.79 \pm 0.04}$ & $\mathbf{0.78 \pm 0.01}$ & $\mathbf{0.960 \pm 0.003}$  \\

\end{tabular}
\end{center}
\end{table}

\begin{figure}[ht]
    \centering
    \includegraphics[width=0.9\textwidth, page=3, trim=0 3.2cm 0 0, clip]{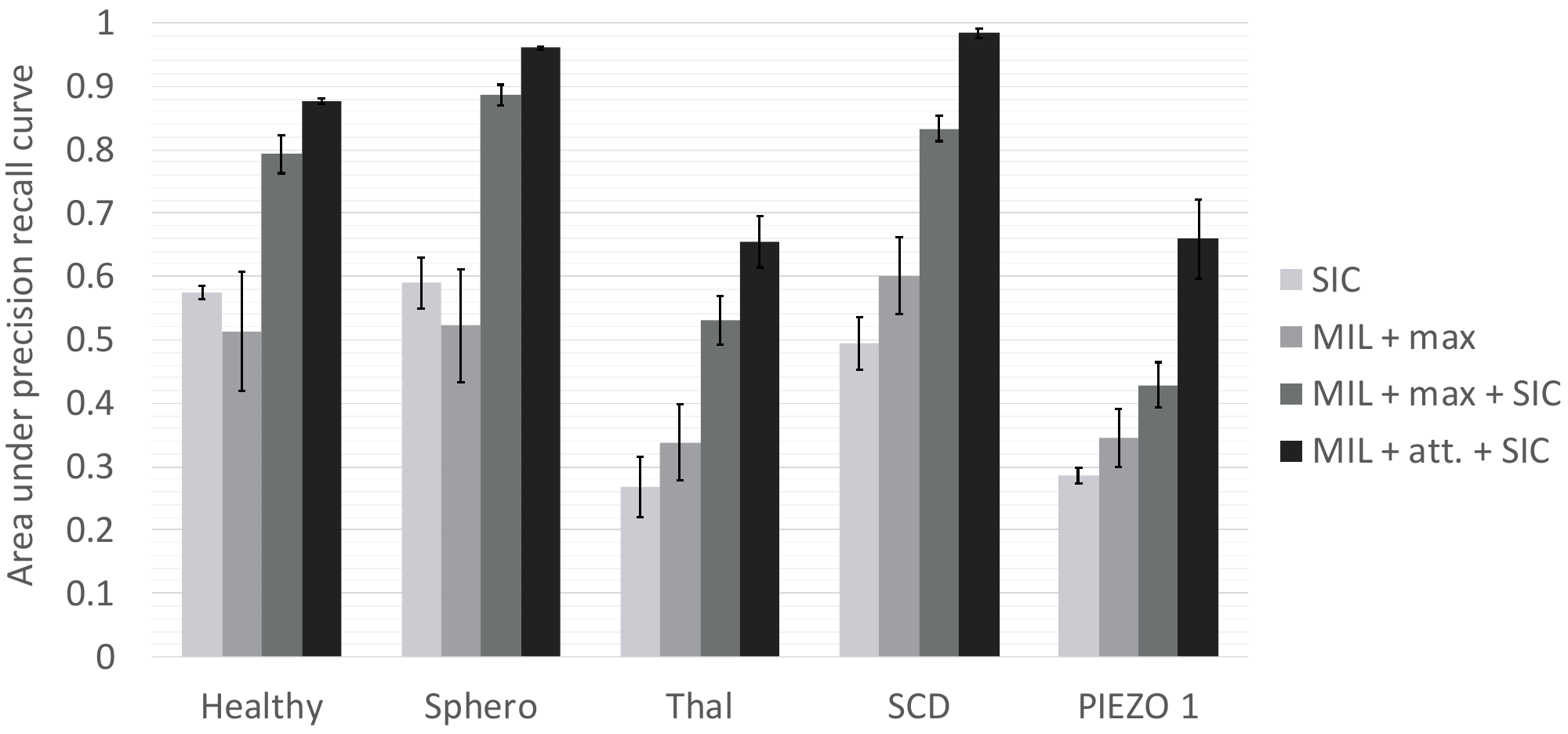}
    \caption{Area under precision recall curve for all experiments and every class is demonstrated. We show mean and standard deviation of five runs.}
    \label{prfigure}
\end{figure}

\begin{figure}
    \centering
    \includegraphics[width=\textwidth,page=4,trim=0 11.5cm 0 0,clip]{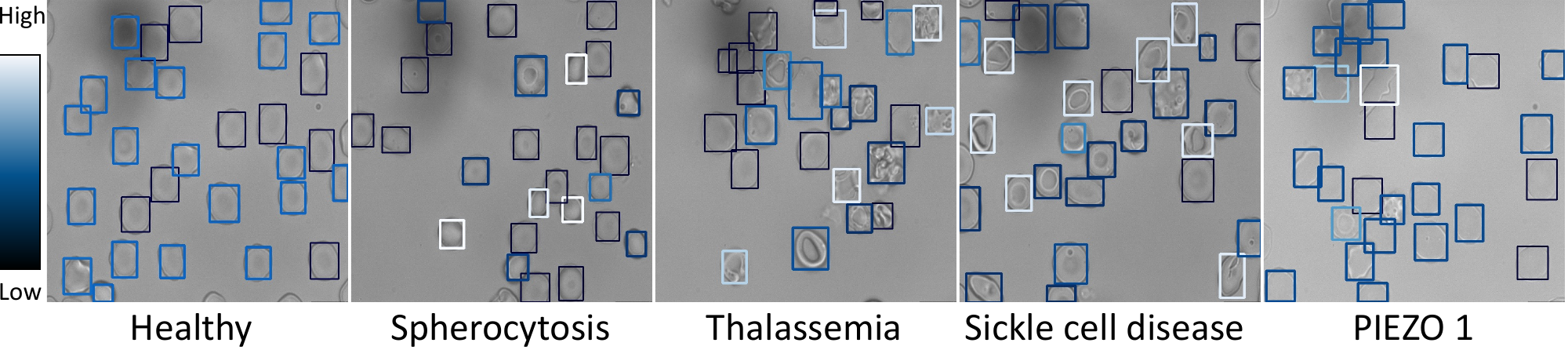}
    \caption{Exemplary whole slide images with attention values, demonstrated with colored bounding boxes. White has the highest attention score while blue and dark blue are the lowest.}
    \label{slideattention}
\end{figure}

\subsection{Ablation study}

All of the experiments were run for five times and the average metric with standard deviation is reported. For each of the experiments we report the accuracy, weighted F1 score and area under the ROC curve in Table \ref{tableresults}. Additionally, in Fig. \ref{prfigure}, the area under precision recall curve for all five classes is reported.

%
%
%
%

\begin{figure}[ht]
    \centering
    \includegraphics[width=\textwidth,page=5, trim=0 5.2cm 0 0,clip]{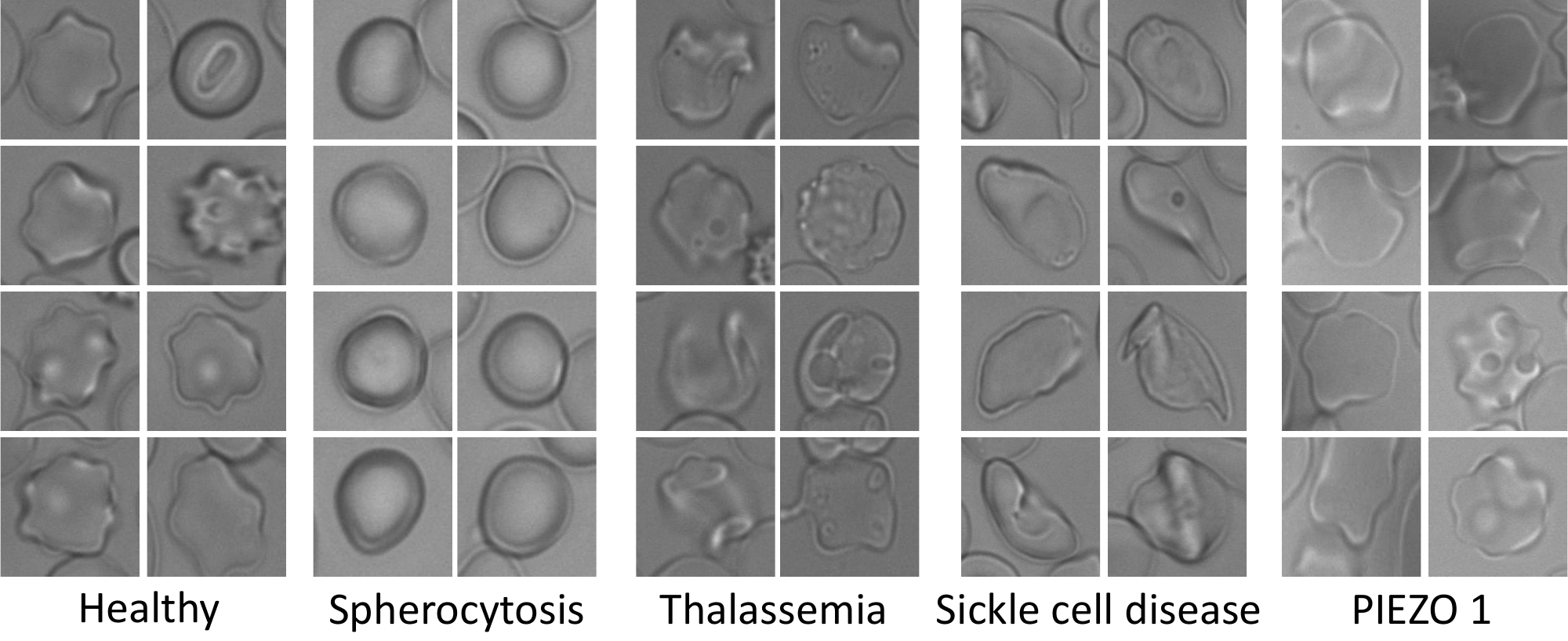}
    \caption{Eight exemplary single cells per class with highest attention.}
    \label{singlecell}
\end{figure}

\subsection{Qualitative assessment and interpretability}
The attention mechanism allows us to have a qualitative evaluation of the model. 
Showing the cells contributing most to the classification can be beneficial for clinical adaptability of the model as it provides the experts with some explanation of the decisions made by the neural network. 
If cells receiving a high attention are known to be important for a specific morphological disorder, not only the model has learned them in an unsupervised manner but this also proves that the model actually knows what is relevant and what is not. 
Figure \ref{slideattention} shows cell attention in an exemplary image from samples belonging to each of the five classes.

Further, we extracted the top eight cells of a sample from every class in the dataset having the highest attention (Fig. \ref{singlecell}).
Cells are clearly morphologically different.
It is interesting to note that for the healthy control class, cells that look a little bit odd received highest attentions as they might be flags for some disorders, although in the end they are not different enough to make the whole sample considered as a disorder. Note that the attention of healthy cells is generally lower than cells from disease samples (Fig. \ref{slideattention}).

\section{Conclusion}
 Our proposed approach based on MIL improves the performance of classification of the genetic blood disorders. The attention mechanism is effective both in terms of accuracy and interpretability of classification. 
 The model automatically learns about diagnostic cells in the samples giving them a high attention. These results are promising and have great potential for decision support and clinical applications in terms of diagnosing blood diseases and training of medical students.

 Possible future works can include uncertainty estimation of the classification \cite{wang2020ud}, including an active learning framework  \cite{settles2008multiple} and using extra features in the attention.
 Additionally, detailed analysis on the shape of the cells receiving high attention might be informative about the underlying pathological mechanisms and severity of the disease manifestation. This might allow stratifying patients and targeted treatments in terms of personalized medicine, which is especially important  for rare anemias such as spherocytosis and xerocytosis, but also for  hemolytic anemia with poor response to conventional treatment \cite{huisjes2020density}.

\section*{Acknowledgments}
C.M. and A.S. have received funding from the European Research Council (ERC) under the European Union’s Horizon 2020 research and innovation programme (Grant agreement No. 866411). C.M. was supported by the BMBF, grant 01ZX1710A-F (Micmode-I2T). S.A. is supported by the PRIME programme of the German Academic Exchange Service (DAAD) with funds from the German Federal Ministry of Education and Research (BMBF).

%
%
%
%
\bibliographystyle{splncs04}
\bibliography{article}

\end{document}